\documentclass[11pt,a4paper]{article}
\usepackage[hyperref]{acl2019}
\usepackage{times}
\usepackage{latexsym}
\usepackage{todonotes}
\usepackage{url}
\usepackage{booktabs}
\usepackage{basecommon}
\usepackage{caption}
\aclfinalcopy 
\def\aclpaperid{2742} 

\title{Label-Agnostic Sequence Labeling by Copying Nearest Neighbors}

\author{Sam Wiseman \qquad Karl Stratos \\
  Toyota Technological Institute at Chicago \\
  Chicago, IL, USA \\
{\tt \{swiseman,stratos\}@ttic.edu} }

\date{}

\begin{document}

\maketitle
\begin{abstract}
Retrieve-and-edit based approaches to structured prediction, where structures associated with retrieved neighbors are edited to form new structures, have recently attracted increased interest. However, much recent work merely conditions on retrieved structures (e.g., in a sequence-to-sequence framework), rather than explicitly manipulating them. We show we can perform accurate sequence labeling by explicitly (and only) copying labels from retrieved neighbors. Moreover, because this copying is label-agnostic, we can achieve impressive performance when transferring to new sequence-labeling tasks without retraining. We additionally consider a dynamic programming approach to sequence labeling in the presence of retrieved neighbors, which allows for controlling the number of distinct (copied) segments used to form a prediction, and leads to both more interpretable and accurate predictions.
\end{abstract}

\section{Introduction}

Retrieve-and-edit style structured prediction, where a model retrieves a set of labeled nearest 
neighbors from the training data and 
conditions on them to generate the target structure, is a promising approach that has recently received renewed interest
\citep{hashimoto2018retrieve,guu2018generating,gu2018search,weston2018retrieve}. 
This approach captures the intuition that while generating a highly complex structure from scratch may be difficult, editing a sufficiently similar structure or set of structures may be easier. 

Recent work in this area primarily uses the nearest neighbors and their labels simply as an additional context for a sequence-to-sequence style model to condition on. 
While effective, these models may not explicitly capture the discrete operations (like copying) that allow for the neighbors to be edited into the target structure, making interpreting the behavior of the model difficult. Moreover, since many retrieve-and-edit style models condition on dataset-specific labels directly, they may not easily allow for transfer learning and in particular to porting a trained model to a new task with different labels. 

We address these limitations in the context of sequence labeling by developing a simple label-agnostic model that explicitly models copying token-level labels from retrieved neighbors.
Since the model is not a function of the labels themselves but only of a learned notion of 
similarity between an input and retrieved neighbor inputs, it can be effortlessly ported to a task with different labels, without any retraining. Such a model can also take advantage of recent advances in representation learning, such as BERT~\citep{devlin2018bert}, in defining this similarity. 

We evaluate the proposed approach on standard sequence labeling tasks, and show it is competitive with label-dependent approaches when trained on the same data, but substantially outperforms strong baselines when it comes to transfer applications, such as when training with coarse labels and testing with fine-grained labels.

Finally, we propose a dynamic programming based approach to sequence labeling in the presence of retrieved neighbors, which allows for trading off token-level prediction confidence with trying to minimize the number of distinct \textit{segments} in the overall prediction that are taken from neighbors. We find that such an approach allows us to both increase the interpretability of our predictions as well as their accuracy.

\section{Related Work}
Nearest neighbor based structured prediction (also referred to as instance- or memory-based learning) has a long history in machine learning and NLP, with early successes dating back at least to the taggers of Daelemans~\citep{daelemans1993memory,daelemans1996mbt} and the syntactic disambiguation system of \citet{cardie1994domain}. Similarly motivated approaches remain popular for computer vision tasks, especially when it is impractical to learn a parametric labeling function~\citep{shakhnarovich2006nearest, schroff2015facenet}.

More recently, there has been renewed interest in explicitly conditioning structured predictions on retrieved neighbors, especially in the context of language generation~\citep{hashimoto2018retrieve,guu2018generating,gu2018search,weston2018retrieve}, although much of this work uses neighbors as extra conditioning information within a sequence-to-sequence framework~\citep{sutskever2014sequence}, rather than making discrete edits to neighbors in forming new predictions. 

Retrieval-based approaches to structured prediction appear particularly compelling now with the recent successes in contextualized word embedding~\citep{mccann2017learned,peters2018deep,radford2018improving,devlin2018bert}, which should allow for expressive representations of sentences and phrases, which in turn allow for better retrieval of neighbors for structured prediction.



Finally, we note that there is a long history of transfer-learning based approaches to sequence labeling~\citep[\textit{inter alia}]{ando2005framework,daume2007frustratingly,schnabel2014flors,zirikly2015cross,peng2016improving,yang2017transfer,rodriguez2018transfer}, though it generally requires some retraining. There has, however, been recent work in zero-shot transfer for sequence labeling problems with binary token-labels~\citep{rei2018zero}.

\begin{figure}[t!]
    \centering
\includegraphics[scale=0.44]{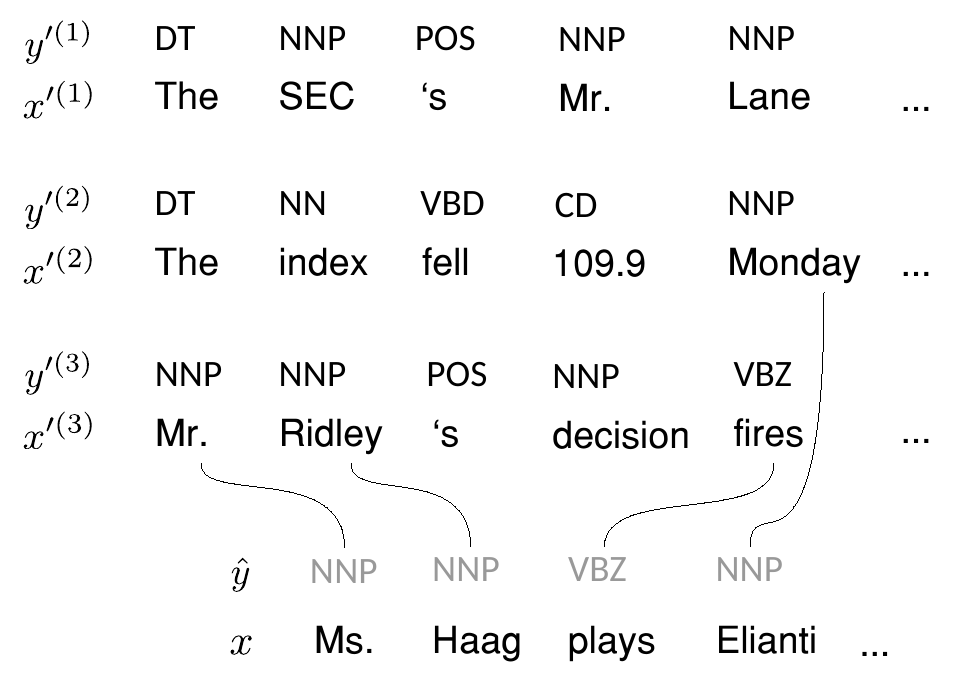}
    \caption{A visualization of POS tagging an input sentence $x$ (bottom) by copying token-labels from $M=3$ retrieved sentences $x'^{(m)}$ for which we know the true corresponding label sequences $y'^{(m)}$; see the text for details.}
    \label{fig:example}
\end{figure}

\section{Nearest Neighbor Based Sequence Labeling}
\label{sec:models}
While nearest-neighbor style approaches are compelling for many structured prediction problems, we will limit ourselves here to sequence-labeling problems, such as part-of-speech (POS) tagging or named-entity recognition (NER), where we are given a $T$-length sequence $x = x_{1:T}$ (which we will assume to be a sentence), and we must predict a corresponding $T$-length sequence of labels $\hat{y} = \hat{y}_{1:T}$ for $x$. We will assume that for any given task there are $Z$ distinct labels, and denote $x$'s true but unknown labeling as $y \, {=} \, y_{1:T} \, {\in} \, \{1, \ldots, Z\}^T$.

Sequence-labeling is particularly convenient for nearest-neighbor based approaches, since a prediction $\hat{y}$ can be formed by simply concatenating labels extracted from the label-sequences associated with neighbors. In particular, we will assume we have access to a database $\mcD = \{x'^{(m)}, y'^{(m)}\}_{m=1}^M$ of $M$ retrieved sentences $x'^{(m)}$ and their corresponding true label-sequences $y'^{(m)}$. We will predict a labeling $\hat{y}$ for $x$ by considering each token $x_t$, selecting a labeled token $x'^{(m)}_k$ from $\mcD$, and then setting $\hat{y}_t = y'^{(m)}_k$.\footnote{More precisely, we will set $\hat{y}_t$ to be an instance of the label \textit{type} of which $y'^{(m)}_k$ is a label token; this distinction between label types and tokens can make the exposition unnecessarily obscure, and so we avoid it when possible.} 

\subsection{A Token-Level Model}
We consider a very simple token-level model for this label-agnostic copying, where the probability that $x$'s $t$'th label $y_t$ is equal to $y'^{(m)}_k$ --- the $k$'th label token of sequence $x'^{(m)}$ --- simply depends on the similarity between $x_t$ and $x'^{(m)}_k$, and is independent of the surrounding labels, conditioned on $x$ and $\mcD$.\footnote{While recent sequence labeling models~\citep{ma2016end,lample2016neural}, often model inter-label dependence with a first-order CRF~\citep{lafferty01conditional}, \citet{devlin2018bert} have recently shown that excellent performance can be obtained by modeling labels as being conditionally independent given a sufficiently expressive representation of $x$.} In particular, we define
\begin{align} \label{eq:model}
    p(y_t \, {=} \, y'^{(m)}_k \given x, \mcD) \propto \exp(\boldx_t^{\trans} \boldx'^{(m)}_k),
\end{align}
where the above probability is normalized over all label tokens of all label-sequences in $\mcD$.
Above, $\boldx_t$ and $\boldx'^{(m)}_k$ (both in $\reals^D$) represent the contextual word embeddings of the $t$'th token in $x$ and the $k$'th token in $x'^{(m)}$, respectively, as obtained by running a deep sequence-model over $x$ and over $x'^{(m)}$. In all experiments we use BERT~\citep{devlin2018bert}, a model based on the Transformer architecture~\citep{vaswani2017attention}, to obtain contextual word embeddings.

We fine-tune these contextual word embeddings by maximizing a latent-variable style probabilistic objective
\begin{align} \label{eq:obj}
   \sum_{t=1}^T \ln \sum_{m=1}^M \sum_{k: \, y'^{(m)}_k \, {=} \, y_t} p(y_t = y'^{(m)}_k \given x, \mcD),
\end{align}
where we sum over all individual label tokens in $\mcD$ that match $y_t$.

At test time, we predict $\hat{y}_t$ to be the label \textit{type} with maximal marginal probability. That is, we set $\hat{y}_t$ to be $\argmax_z \sum_{m=1}^M \sum_{k: \, y'^{(m)}_k = z} p(y_t \, {=} \, y'^{(m)}_k \given x, \mcD)$, where $z$ ranges over the label types (e.g., POS or named entity tags) present in $\mcD$. As noted in the introduction, predicting labels in this way allows for the prediction of any label type present in the database $\mcD$ used at test time, and so we can easily predict label types unseen at training time without any additional retraining.


\section{Data and Methods}
Our main experiments seek to determine both whether the label-agnostic copy-based approach introduced above results in competitive sequence-labeling performance on standard metrics, as well as whether this approach gives rise to better transfer. Accordingly, our first set of experiments consider several standard sequence-labeling tasks and datasets, namely, POS tagging the Penn Treebank~\citep{marcus1993building} with both the standard Penn Treebank POS tags and Universal POS tags~\citep{petrov2012universal,nivre2016universal}, and the CoNLL 2003 NER task~\citep{sang2000introduction,sang2003introduction}. We compare with the sequence-labeling performance of BERT~\citep{devlin2018bert}, which we take to be the current state of the art. We use the standard dataset-splits and evaluations for all tasks, and BIO encoding for all segment-level tagging tasks.

We evaluate transfer performance by training on one dataset and evaluating on another, without any retraining. In particular, we consider three transfer scenarios: training with Universal POS Tags on the Penn Treebank and then predicting the standard, fine-grained POS tags, training on the CoNLL 2003 NER data and predicting on the fine-grained OntoNotes NER data~\citep{hovy2006ontonotes} using the setup of \citet{strubell2017fast}, and finally training on the CoNLL 2003 chunking data and predicting on the CoNLL 2003 NER data. We again compare with a BERT baseline, where labels from the original task are deterministically mapped to the most frequent label on the new task with which they coincide.\footnote{For the Chunk $\rightarrow$ NER task, this results in mapping all tags to `O', so we instead use the more favorable mapping of NPs to PERSON tags.} 

Our nearest-neighbor based models were fine-tuned by retrieving the 50 nearest neighbors of each sentence in a mini-batch of either size 16 or 20, and maximizing the objective~\eqref{eq:obj} above. For training, nearest neighbors were determined based on cosine-similarity between the averaged top-level (non-fine-tuned) BERT token embeddings of each sentence. In order to make training more efficient, gradients were calculated only with respect to the input sentence embeddings (i.e., the $\boldx_t$ in \eqref{eq:model}) and not the embeddings $\boldx'^{(m)}_k$ of the tokens in $\mcD$. At test time, 100 nearest neighbors were retrieved for each sentence to be labeled using the fine-tuned embeddings. 

The baseline BERT models were fine-tuned using the publicly available \texttt{huggingface} BERT implementation,\footnote{\url{https://github.com/huggingface/pytorch-pretrained-BERT}} and the ``base'' weights made available by the BERT authors~\citep{devlin2018bert}. We made word-level predictions based on the embedding of the first tokenized word-piece associated with a word (as \citet{devlin2018bert} do), and ADAM~\citep{kingma2014adam} was used to fine-tune all models. Hyperparameters were chosen using a random search over learning rate, batch size, and number of epochs. Code for duplicating all models and experiments is available at \url{https://github.com/swiseman/neighbor-tagging}.

\section{Main Results}
The results of our experiments on standard sequence labeling tasks are in Table~\ref{tab:stdresults}. We first note that all results are quite good, and are competitive with the state of the art. The label-agnostic model tends to underperform the standard fine-tuned BERT model only very slightly, though consistently, and is typically within several tenths of a point in performance. 

\begin{table}[t!]
\small
\centering
\begin{tabular}{lcc}
\toprule
NER & Dev. F$_1$ & Test F$_1$  \\
\midrule
BERT & 95.14 & 90.76 \\
NN  & 94.48 & 89.94 \\
\midrule
POS & Dev. Acc. & Test Acc.  \\
\midrule
BERT & 97.56 & 97.91 \\
NN  & 97.33 & 97.64 \\
\midrule
U-POS & Dev. Acc. & Test Acc.  \\
\midrule
BERT & 98.34 & 98.62 \\
NN  & 98.08 & 98.36 \\
\bottomrule
\end{tabular}
\caption{A comparison of fine-tuned BERT and our nearest-neighbor (NN) based approach on standard sequence labeling tasks. From top-to-bottom, NER performance on the CoNLL 2003 data, part-of-speech tagging performance on the Penn Treebank, and universal part-of-speech tagging performance on the Penn Treebank; results use the standard metrics and dataset splits. BERT numbers are from fine-tuning the \texttt{huggingface} BERT implementation, and differ slightly from those in \citet{devlin2018bert}.}
\label{tab:stdresults}
\end{table}

The results of our transfer experiments are in Table~\ref{tab:zeroresults}. We see that in all cases the label-agnostic model outperforms standard fine-tuned BERT, often significantly. In particular, we note that when going from universal POS tags to standard POS tags, the fine-tuned label-agnostic model manages to outperform the standard most-frequent-tag-per-word baseline, which itself obtains slightly less than 92\% accuracy. The most dramatic increase in performance, of course, occurs on the Chunking to NER task, where the label-agnostic model is successfully able to use chunking-based training information in copying labels, whereas the parametric fine-tuned BERT model can at best attempt to map NP-chunks to PERSON labels (the most frequent named entity in the dataset). 

In order to check that the increase in performance is not just due to BERT's pretraining, Table~\ref{tab:zeroresults} also shows the results of the label-agnostic model without fine-tuning (as indicated by ``no FT'' in the table). In all cases, this leads to a decrease in performance.



\begin{table}[t!]
\small
\centering
\begin{tabular}{lll}
\toprule
CoNLL $\rightarrow$ Onto NER & Dev. F$_1$ & Test F$_1$  \\
\midrule
BERT & 58.41 & 58.05 \\
NN & 62.17 & 62.33 \\
NN (no FT) & 54.29 & 55.35 \\
\midrule
U-POS $\rightarrow$ POS & Dev. Acc. & Test Acc.  \\
\midrule
BERT & 61.78 & 59.86 \\
NN & 96.70  & 96.98  \\
NN (no FT) & 87.44  & 87.13 \\
\midrule
Chunk $\rightarrow$ NER & Dev. F$_1$ & Test F$_1$ \\
\midrule
BERT & \,9.55 & \,8.03 \\
NN & 78.05 & 71.74 \\
NN (no FT) & 75.21 & 67.19 \\
\bottomrule
\end{tabular}
\caption{From top-to-bottom, transfer performance of models trained on the CoNLL 2003 data and applied to the fine-grained OntoNotes NER task, on PTB with universal part-of-speech tags and applied to PTB with standard part-of-speech tags, and on the CoNLL 2003 chunking data and applied to the CoNLL 2003 NER task. Above, ``no FT'' indicates the model was not fine tuned even on the original task.}
\label{tab:zeroresults}
\end{table}

\section{Encouraging Contiguous Copies}
Although we model token-level label copying, at test time each $\hat{y}_t$ is predicted by selecting the label type with highest marginal probability, without any attempt to ensure that the resulting sequence $\hat{y}$ resembles one or a few of the labeled neighbors $y'^{(m)}$. 
In this section we therefore consider a decoding approach that allows for controlling the trade-off between prediction confidence and minimizing the number of \textit{distinct} segments in $\hat{y}$ that represent direct (segment-level) copies from some neighbor, in the hope that having fewer distinct copied segments in our predictions might make them more interpretable or accurate. We emphasize that the following decoding approach is in fact applicable even to standard sequence labeling models (i.e., non-nearest-neighbor based models), as long as neighbors can be retrieved at test time.

\begin{figure*}[t!]
    \centering
    \fbox{\includegraphics[scale=0.41]{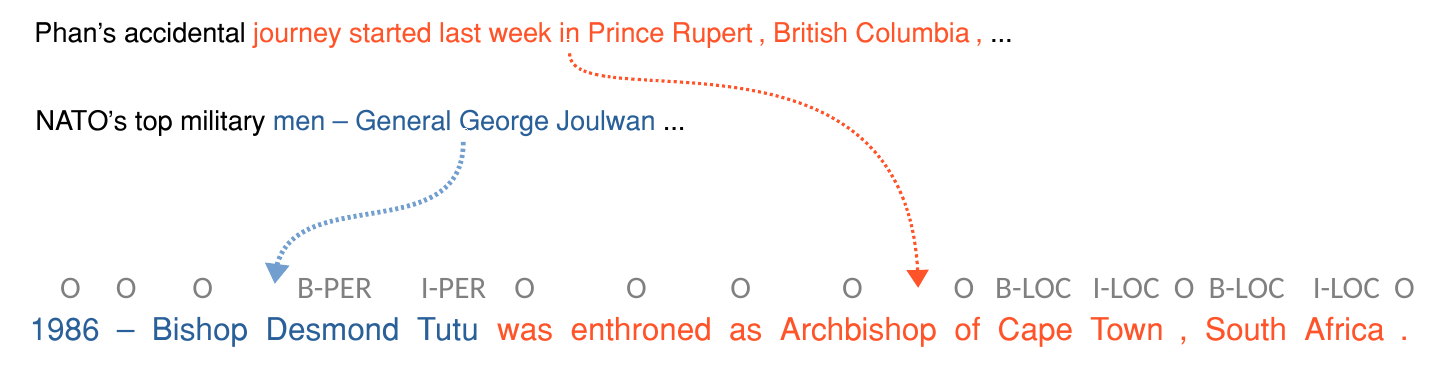}}
    \fbox{\includegraphics[scale=0.41]{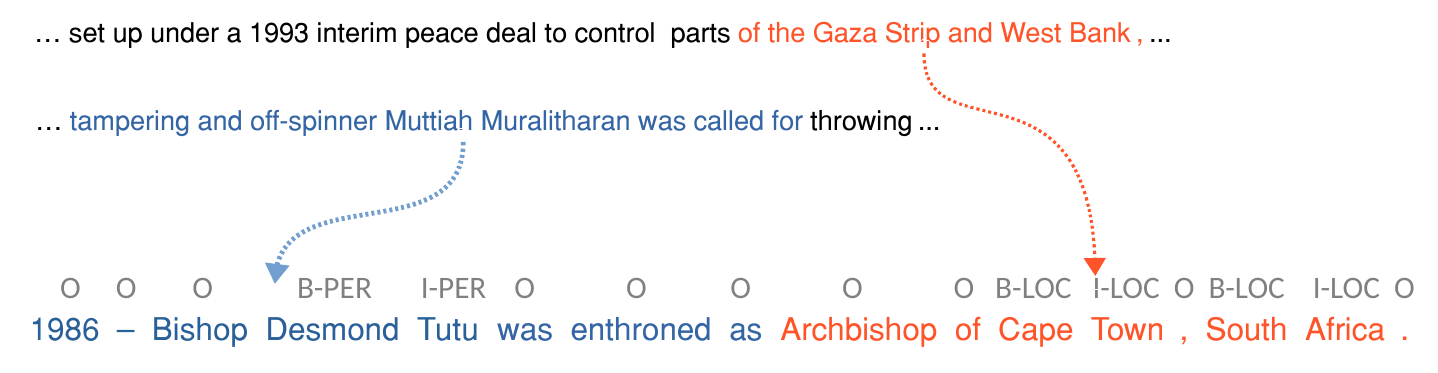}}
    \caption{A CoNLL 2003 NER development example, which can be labeled with only two distinct segments. We show the segments used by a model trained on the NER data (top), and by a model trained on the CoNLL chunking data (bottom).}    
    \label{fig:examples}
\end{figure*}

To begin with a simple case, suppose we already know the true labels $y$ for a sequence $x$, and are simply interested in being able to reconstruct $y$ by concatenating as few segments $y'_{i:j}$ that appear in some $y'^{(m)} \, {\in} \, \mcD$ as possible. 
More precisely, define the set $\mcZ_{\mcD}$ to contain all the unique label \textit{type} sequences appearing as a subsequence of some sequence $y'^{(m)} \, {\in} \, \mcD$. 
Then, if we're willing to tolerate some errors in reconstructing $y$, we can use a dynamic program to minimize the number of mislabelings in our now ``prediction'' $\hat{y}$, plus the number of distinct segments used in forming $\hat{y}$ multiplied by a constant $c$, as follows:
\begin{align*}
    J(t) = \hspace*{-4mm} \min_{\substack{1 \leq k \leq t \\ z {\in} \mcZ_{\mcD}: |z| = k}} \hspace*{-3mm} J(t{-}k) + c + \hspace*{-0.5mm} \sum_{j=1}^{k}\mathbf{1}[y_{t{-}k{+}j} \, {\neq} \,z_{j}],
\end{align*}
where $J(0) \, {=} \, 0$ is the base case and $|z|$ is the length of sequence $z$. Note that greedily selecting sequences that minimize mislabelings may result in using more segments, and thus a higher $J$.

In the case where we do not already know $y$, but wish to predict it, we might consider a modification of the above, which tries to minimize $c$ times the number of distinct segments used in forming $\hat{y}$ plus the expected number of mislabelings:
\begin{align*}
    J(t) = \hspace*{-4mm} \min_{\substack{1 \leq k \leq t \\ z {\in} \mcZ_\mcD: |z| = k}} \hspace*{-3mm} &\big[J(t{-}k) + c \\
    &+ \hspace*{-0.5mm} \sum_{j=1}^{k}1 \, {-} \, p(y_{t{-}k{+}j} \, {=} \, z_j\given x, \mcD)\big],
\end{align*}
where we have used the linearity of expectation.
Note that to use such a dynamic program to predict $\hat{y}$ we only need an estimate of $p(y_{t{-}k{+}j} \, {=} \, z_j\given x, \mcD)$, which we can obtain as in Section~\ref{sec:models} (or from a more conventional model).

\begin{figure}[t!]
    \centering
\includegraphics[scale=0.48]{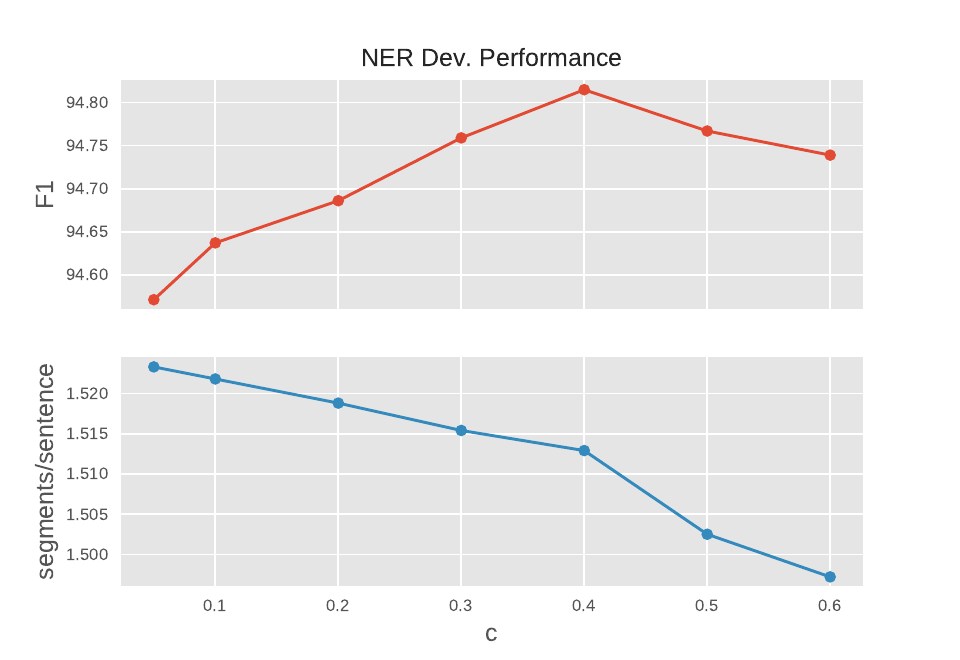} \\
\includegraphics[scale=0.48]{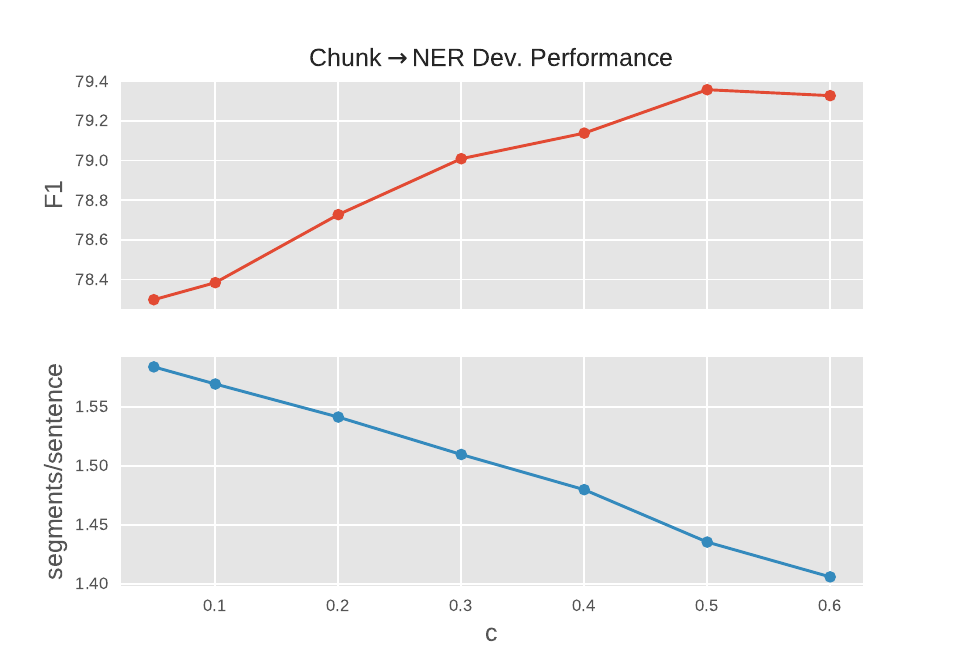}
    \caption{F$_1$ performance (top, orange suplots) on the CoNLL 2003 NER development data and the average number of distinct segments per predicted labeling (bottom, blue subplots) as the $c$ parameter is varied, when the model is trained either (top) on the standard training set or (bottom) on the CoNLL chunking data (i.e., transfer performance).}
    \label{fig:plot}
\end{figure}

In Figure~\ref{fig:plot} we plot both the F$_1$ score and the average number of distinct segments used in predicting each $\hat{y}$ against the $c$ parameter from the dynamic program above, for the CoNLL 2003 NER development data in both the standard and transfer settings. First we note that we are able to obtain excellent performance with only about 1.5 distinct segments per prediction, on average; see Figure~\ref{fig:examples} for examples. Interestingly, we also find that using a higher $c$ (leading to fewer distinct segments) can in fact improve performance. Indeed, taking the best values of $c$ from Figure~\ref{fig:plot} (0.4 in the standard setting and 0.5 in the transfer setting), we are able to improve our performance on the test set from 89.94 to 90.20 in the standard setting and from 71.74 to 73.61 in the transfer setting, respectively; see Tables~\ref{tab:stdresults} and ~\ref{tab:zeroresults}.

\section{Conclusion}
We have proposed a simple label-agnostic sequence-labeling model, which performs nearly as well as a standard sequence labeler, but improves on transfer tasks. We have also proposed an approach to sequence label prediction in the presence of retrieved neighbors, which allows for discouraging the use of many distinct segments in a labeling. Future work will consider problems where more challenging forms of neighbor manipulation are necessary for prediction.

\bibliography{acl2019}
\bibliographystyle{acl_natbib}

\end{document}